\documentclass{article}

\usepackage{PRIMEarxiv}

\usepackage{amssymb}
\usepackage{amsmath}
\usepackage{float}
\usepackage{xcolor}
\usepackage{xspace}
\usepackage{booktabs}
\usepackage{makecell}
\usepackage{multirow}
\usepackage{verbatim}
\usepackage{subcaption}
\usepackage[colorlinks=true,linkcolor=blue]{hyperref}
\usepackage[linesnumbered,ruled,vlined]{algorithm2e}
\usepackage{enumerate}
\usepackage[shortlabels]{enumitem}

\usepackage{xurl}
\usepackage{helvet}
\usepackage{etoolbox}
\usepackage{graphicx}
\usepackage{titlesec}
\usepackage{caption}
\usepackage{scrextend}
\usepackage{amsfonts}
\usepackage{bm}
\usepackage{amsmath}
\usepackage{soul}

\usepackage{tikz}
\relax
\usepackage[edges]{forest}
\usetikzlibrary{shadows.blur}

\newcommand\revision[1]{{\color{black}#1}}

\title{General Purpose Artificial Intelligence Systems (GPAIS): Properties, Definition, Taxonomy, Open Challenges and Implications
}

\author{
  Isaac Triguero$^{1,4}$, Daniel Molina$^{1}$, Javier Poyatos$^{1}$, Javier Del Ser$^{2,3}$, Francisco Herrera$^{1}$\\
  $^1$ Department of Computer Science and Artificial Intelligence, Andalusian Research Institute \\ in Data Science and Computational Intelligence (DaSCI), University of Granada, Granada, 18071, Spain \\
  $^2$ TECNALIA, Basque Research Technology Alliance (BRTA), Derio, 48160, Spain\\  
  $^3$ University of the Basque Country (UPV/EHU), Bilbao, 48013, Spain \\
  $^4$ School of Computer Science, University of Nottingham, Nottingham, NG8 1BB, United Kingdom\\
  \texttt{Corresponding author: triguero@decsai.ugr.es} \\
}

\begin{document}
\maketitle

\begin{abstract}
Most applications of Artificial Intelligence (AI) are designed for a confined and specific task. However, there are many scenarios that call for a more general AI, capable of solving a wide array of tasks without being specifically designed for them. The term General Purpose Artificial Intelligence Systems (GPAIS) has been defined to refer to these AI systems. To date, the possibility of an Artificial General Intelligence, powerful enough to perform any intellectual task as if it were human, or even improve it, has remained an aspiration, fiction, and considered a risk for our society. Whilst we might still be far from achieving that, GPAIS is a reality and sitting at the forefront of AI research.

This work discusses existing definitions for GPAIS and proposes a new definition that allows for a gradual differentiation among types of GPAIS according to their properties and limitations. We distinguish between closed-world and open-world GPAIS, characterising their degree of autonomy and ability based on several factors such as adaptation to new tasks, competence in domains not intentionally trained for, ability to learn from few data, or proactive acknowledgment of their own limitations. We then propose a taxonomy of approaches to realise GPAIS, describing research trends such as the use of AI techniques to improve another AI (commonly referred to as \emph{AI-powered AI}) or (single) foundation models. As a prime example, we delve into generative AI (GenAI), aligning them with the terms and concepts presented in the taxonomy. \revision{Similarly, we explore the challenges and prospects of multi-modality, which involves fusing various types of data sources to expand the capabilities of GPAIS.} Through the proposed definition and taxonomy, our aim is to facilitate research collaboration across different areas that are tackling general purpose tasks, as they share many common aspects. \revision{Finally, with the goal of providing a holistic view of GPAIS, we discuss the current state of GPAIS, its prospects, implications for our society, and the need for regulation and governance of GPAIS to ensure their responsible and trustworthy development.}
\end{abstract}


\section{Introduction}\label{sec:introduction}
The recent advances in Large Language Models (LLMs) \cite{Mikolov2013}, such as ChatGPT, may be perceived as a step towards getting to Artificial General Intelligence (AGI) \cite{Grudin2019}, in which a machine could think for itself, matching or exceeding human capabilities. This has created a lot of hype and fears about the development of AI \cite{Kaplan201915}. While these models seem to be able to perform some tasks which they were not directly trained for, it is yet unclear whether such models do display emergent abilities on unseen tasks \cite{wei2022emergent, schaeffer2023emergent, Bubeck2023}. The intelligence of current AI systems significantly differs from human intelligence \cite{Korteling2021}, as they may be lacking other abilities (e.g. complex reasoning, consciousness, or initiative, among others), but they possess others (e.g. assembling patterns) that may be great to supporting humans in complex tasks. In the literature, the term AGI is often used but ill-defined \cite{gutierrez2023proposal}. This has made it difficult to say how far we are currently from realising AGI, if at all possible \cite{Shevlin2019, Fjelland2020}.

Actually, thus far, the most extensive use of AI models today falls within those that are capable of performing confined and specific tasks (also known as narrow or fixed-purpose AI). Typically, these models are manually designed by experts, following numerous steps from data collection, through data modelling, to deployment, in a pipeline referred to as the Machine Learning (ML) Lifecycle \cite{Ashmore2021}. These models are necessary and are being established in many areas of our society with a high volume of data, where designing and exploiting AI algorithms by experts is doable. Unfortunately, these models usually lack the generalisation abilities to perform well on unseen tasks, so that, their practical application remains bound to the tasks it was trained for. Nevertheless, numerous General-Purpose AI systems (GPAIS\footnote{Throughout the document, GPAIS may represent either a singular system or multiple systems depending on the context.}) have recently emerged to solve more than one task and to be able to generalise to unseen tasks with very few data (or even none) and/or adjustments.

While AGI remains an aspiration, the term GPAIS can be regarded as a more modest and realistic expectation of AI, which would not expect other abilities inherent to humans, as mentioned above. In the best-case scenario, GPAIS would also be capable of dealing with tasks without being directly programmed to do so, as soon as it has seen data that could help it solve them. To enable GPAIS, one option is to add a new layer of abstraction that would use an AI algorithm to design or enrich another AI algorithm. In a nutshell, we \emph{construct} or \emph{enhance} AI with an additional AI stage. A common example of this is to make AI learn to work like an AI expert, determining which algorithms and/or components are most suitable for a given problem \cite{Hutter2019}. Alternatively, we may aim to create a single model capable of exploiting large amounts of data for multiple tasks, as multi-task learning \cite{Zhang2018} does, or include a fine-tuning step to adapt to new tasks, as foundation models implement \cite{Bommasani2021}. \revision{In the literature, the term \textbf{foundation models} is frequently employed interchangeably with GPAIS. Similarly, with the growing prevalence of LLMs, there is a tendency to conflate the concept of generative AI (GenAI) with that of general purpose AI. In this work, we contend that while these terms are integral to GPAIS, they do not encompass their entirety.}

Among others, one may expect a GPAIS to be able to transfer knowledge from similar tasks, learn with as little data as possible, and rapidly adapt to changes and/or new tasks. These properties are typically needed in a wide range of research areas such as AutoML \cite{real2020automl}, few-shot learning \cite{Wang2020}, or continual learning \cite{Parisi2019}, some of which are solved by meta-learning \cite{Lemke2015}, reinforcement learning \cite{Silver2021} or evolutionary computation \cite{Stanley201924}. To regulate and manage the risks of generally capable AIs \cite{Hendrycks2023, Critch2023}, different authors have recently defined what a GPAIS may be \cite{gutierrez2023proposal,Campos2023}. Apart from some level of disagreement across definitions, those works provided high-level definitions without landing their definitions based on the existing research. 

The goal of this paper is to become a primer for researchers and practitioners interested in a holistic vision of GPAIS, mapping out the desiredata of a GPAIS and describing different approaches to build those models. Our analysis tackles the main aspects of GPAIS:

\begin{itemize} [leftmargin=*]

 \item  \textit{Proposing a new definition together with its properties}: First, we will discuss what a GPAIS is, looking in detail at what is considered to be a GPAIS in the literature and how it has evolved, analysing four previous definitions. Then, we articulate the differences between GPAIS, traditional fixed-purpose AI, and AGI to provide a complete definition and categorisation of GPAIS. We will describe the most relevant features of GPAIS, considering different levels of ability and autonomy, and distinguishing between various degrees of GPAIS. Taking into consideration the variable time, we differentiate between closed-world GPAIS and open-world GPAIS.

 \item \textit{Breaking it down onto a taxonomy}: We devise a taxonomy of approaches to building GPAIS, describing some of the key research trends and problems in which GPAIS are being developed. We identify how these approaches may fit within the proposed definition and the capabilities they may provide. This includes GPAIS which use a single model to generalise to many tasks and AI-powered AI approaches.
 
\item \revision{\textit{Linking the proposed taxonomy with trending topics in AI: GenAI and multi-modality}. First, we delve into GenAI \cite{Stokel-Walker2023} as the first type of GPAIS perceived as general intelligence by the public. Next, we talk about the development of GPAIS capable of exploiting (and fusing) multiple types of data sources (e.g. image, text, audio) to learn a more cohesive representation of concepts in the same fashion that humans do.}

 \item \textit{Providing a detailed discussion around GPAIS}:  We debate the state of GPAIS and their challenges, including their limitations and prospects, and explore how some of the approaches identified in the proposed taxonomy could help advance this field. Finally, we reflect on the impact of GPAIS on our society and emphasise the urgent need for trustworthy and sustainable systems together with appropriate regulations that dictate how these models must be audited and accounted for. We claim that an understanding of the design principles behind these techniques might aid in establishing appropriate regulations \revision{and necessary governance for the safe deployment of GPAIS.}

\end{itemize}

\begin{figure}[!h]
    \centering
    \includegraphics[width=0.8\textwidth]{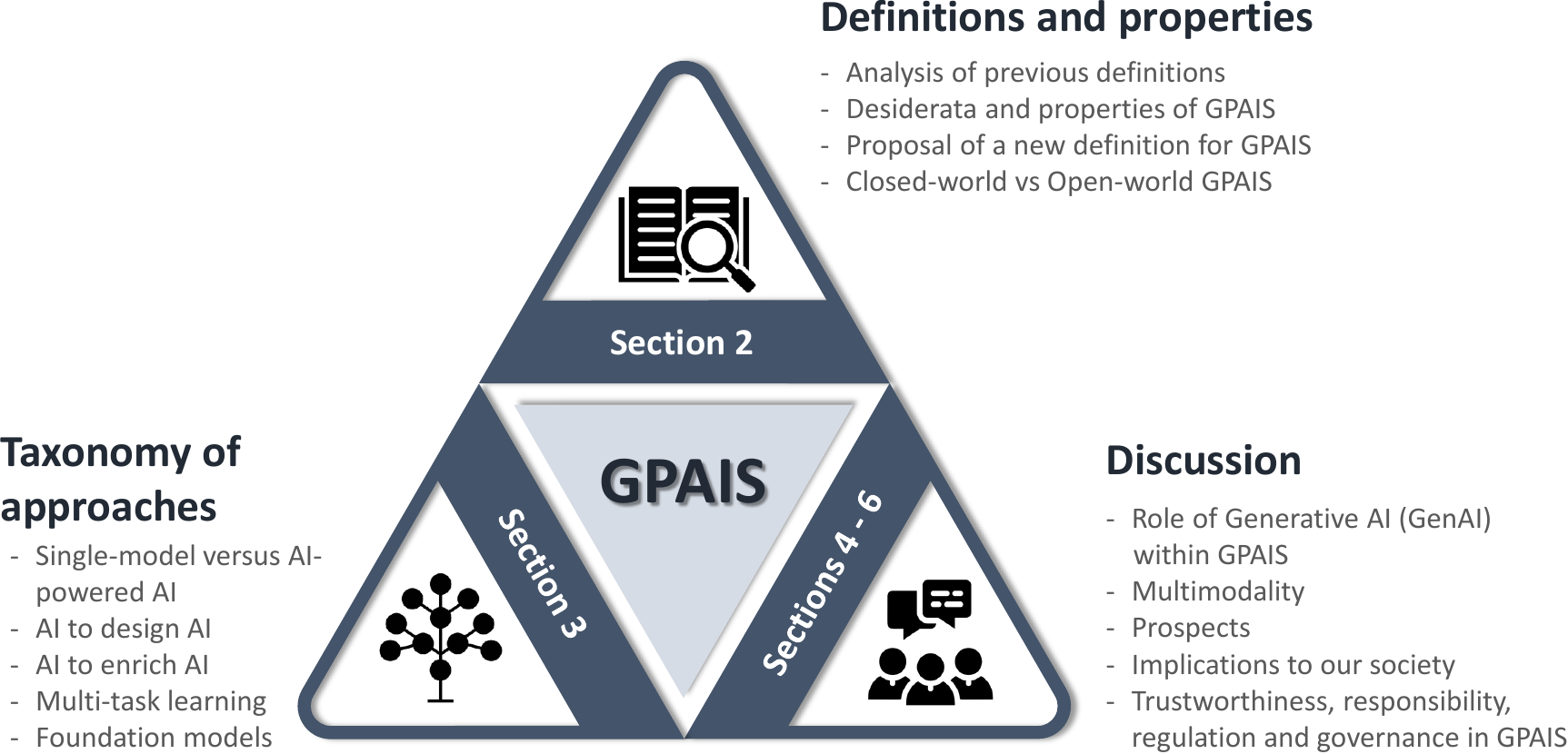}
    \caption{Schematic diagram representing the contributions of this work and their distribution over the sections of the manuscript.}
    \label{fig:avocado0}
\end{figure}

As depicted in Figure \ref{fig:avocado0}, the remainder of this paper is organised as follows. Section \ref{sec:gpais-defs} covers the existing definitions for GPAIS, and provides our definition and categorisation as well as the expected properties of a GPAIS. Section \ref{sec:gpais-taxo} presents the proposed taxonomy of approaches to creating GPAIS. Section \ref{sec:generativemodels} focuses on generative AI models as the most notable example of GPAIS nowadays. \revision{Section \ref{sec:multimodal} elaborates on the use of multi-modality in emerging GPAIS. Section \ref{sec:discussion} explores the present status of GPAIS, their challenges ahead, societal implications and the need for regulation and governance}. Finally, Section \ref{sec:conclusions} concludes the paper with a summary of our contribution.

\section{General-Purpose Artificial Intelligence Systems: Definitions and Properties}\label{sec:gpais-defs}

This section shows and discusses the existing definitions of GPAIS in the literature (Subsection \ref{sec-def}). After that, we establish a new definition and categorise GPAIS together with their key properties (Subsection \ref{sec-defprop}).

\subsection{Definitions related to General Purpose Artificial Intelligence Systems in the literature} 
\label{sec-def}
\newtheorem{definition}{Definition} 

Many have described what a GPAIS may entail, but very few formal definitions exist. We briefly present and analyse recent formal definitions of the concept of GPAIS, or closely related terms, that exist in the literature, most of which revolve around legislation. The risks about the potential of AI, and more recently about GPAIS, have promoted various definitions of what these systems may be and what their capabilities are. Our goal is not to track down all possible definitions of GPAIS, but to identify the most relevant ones and their differences. 

In terms of regulation and legislation, the AI Act of the European Union is the first proposed law on AI in the world. The first article referring to GPAIS defines them as follows:

\begin{definition}[Article 3(1b) AI Act (December 6, 2022) \cite{aiAct2021}]
Intended by the \\ provider to perform generally applicable functions such as image and speech recognition, audio and video generation, pattern detection, question answering, translation and others; a general purpose AI system may be used in a plurality of contexts and be integrated in a plurality of other AI system.
\end{definition}

This definition has been heavily criticised \cite{Hacker2023}, as it is overly inclusive, so that, simple image or speech recognition systems would qualify as GPAIS. Within this context, the Future of Life Institute put forward some recommendations to shape the different articles of the proposed European law. They put forward the following definition of GPAIS:

\begin{definition}[\cite{FLI2022}]
‘General purpose AI system’ means an AI system that is able to perform generally applicable functions such as image/speech recognition, audio/video generation, pattern detection, question answering, translation, etc, and is able to have multiple intended and unintended purposes.
\end{definition}

The most relevant aspects of this definition include the idea of performing many tasks, and the ability of these systems to perform tasks they were not directly trained for. In addition, the Future of Life Institute indicates that GPAIS are characterised by their scale (large memory, abundant data, and powerful hardware) as well as their reliance on transfer learning (applying knowledge from one task to another). Moreover, they also highlight that GPAIS are not limited to a single type of information input, that is, they are multi-modal.

With regard to the use of transfer learning, they refer to the concept of foundation models coined in \cite{Bommasani2021}. Foundation models are based on standard deep learning and transfer learning, and they are defined as a model that ``\textit{is trained on broad data (generally using self-supervision at scale) that can be adapted (e.g., fine-tuned) to a wide range of downstream tasks}''. In this work, we argue that GPAIS goes beyond transfer learning and deep learning, and therefore, foundation models become a subset of GPAIS.

In line with the previous definition, and also in the context of the AI Act, Gutierrez et al. presented in \cite{gutierrez2023proposal} another formal definition for GPAIS. This definition states:

\begin{definition}[\cite{gutierrez2023proposal}]
An AI system that can accomplish or be adapted to accomplish a range of distinct tasks, including some for which it was not intentionally and specifically trained.
\end{definition}

As such, the definition is somewhat equivalent to the previous one, emphasising the possibility of completing tasks outside of those it is specifically trained for. Prior to the definition, the authors claimed that GPAIS resembles the idea of AGI, but their definition does not specify the differences between GPAIS and AGI. The authors also mentioned that GPAIS may have different degrees of autonomy, with or without human intervention. They also mentioned that GPAIS may be trained in different manners (e.g. Gato \cite{reed2022} uses supervised learning, whereas MuZero \cite{Schrittwieser2019} is based on reinforcement learning). 

Expanding on this definition, \cite{Campos2023} recently suggested some edits to Definition 3, yielding:

\begin{definition}[\cite{Campos2023}]
An AI system that can accomplish a range of distinct \textit{valuable} tasks, including some for which it was not specifically trained.
\end{definition}

Specifically, they deleted the words ``be adapted to'' and ``intentionally'', and added the word ``valuable'' to the definition. The nature of those edits is all about the safety and risk-based frameworks that can be used to mitigate the challenges of GPAIS. For example, they considered that the ability of a model to be adapted to accomplish a task provides too many degrees of freedom. The authors exemplified this with the minimal risks associated with a fine-tuned BERT model \cite{Devlin2019}; according to \cite{Campos2023}, a specialised translation model would not easily be able to do any other tasks other than translation. Whilst we agree on the safety aspect covered in this work, narrowing the definition of GPAIS down to the riskiest models would not help advance the development of the different research areas that encompass GPAIS.

Definitions 2 and 3 are very much focused on the adaptation to perform tasks of different types, while Definition 4 is centred on the most generally capable models, neglecting simpler or more primitive methods. Thus, we find Definitions 2 and 3 suitable to generally speak about GPAIS. However, differently from the existing definitions, our focus is not on the legislation of GPAIS, but on the way they are designed, the problems they can solve, and the challenges they are to address to continue progressing. Therefore, the main distinct points to approach a new definition for GPAIS are:

\begin{itemize}
\item Current definitions speak about AGI and GPAIS interchangeably. We consider AGI a much more pretentious goal, in which a machine would have the autonomy of a human being. Conversely, GPAIS is a more practical term that allows us to consider that an AI could fulfil the term at different levels of generality.

\item GPAIS may follow other approaches that are not pure transfer learning only. Thus, we deem foundation methods as an important category within GPAIS, but other strategies exist which do not explicitly use broad data prior to a fine-tuning step. An example of that could be AutoML-zero, which aims to automatically discover novel AI algorithms using a set of tasks without following the pre-training preceded by fine-tuning approach of foundation models. 
\item We categorise GPAIS, so that we may understand their degree of autonomy and expected 
capabilities. We consider that the proposed categorisation helps provide a global view, highlighting the role that different techniques can play for these systems.
\end{itemize}

\subsection{Definition and properties of General Purpose Artificial Intelligence Systems} \label{sec-defprop}

\begin{figure}[htp]
    \centering
    \includegraphics[width=0.8\textwidth]{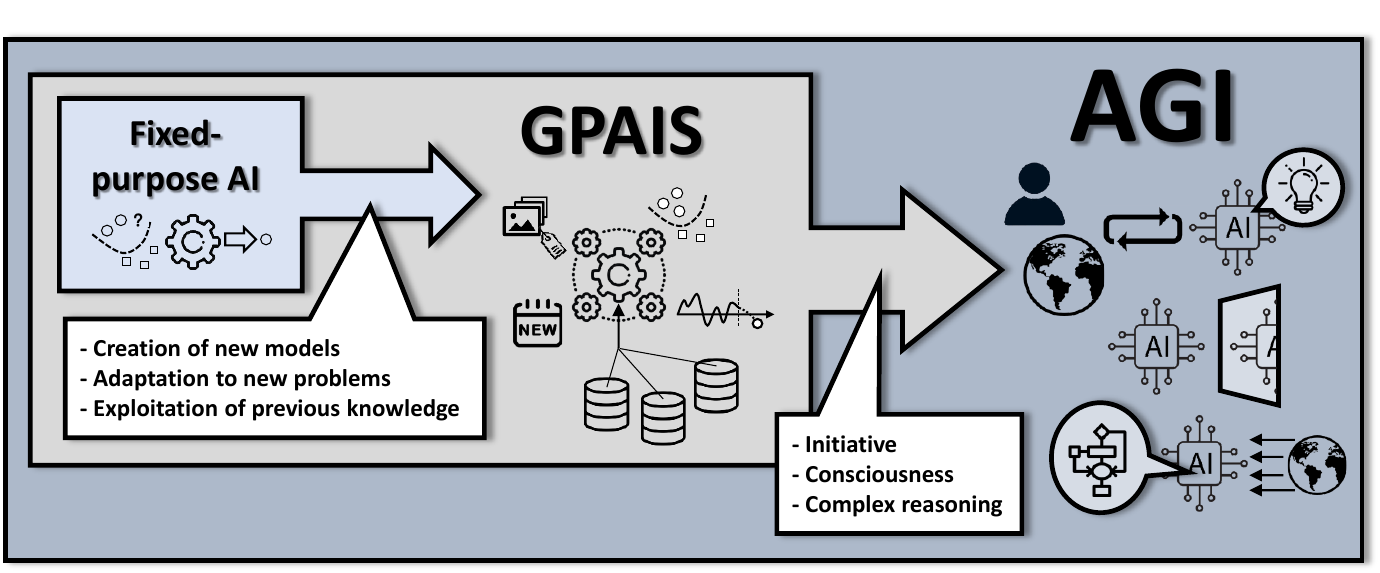}
    \caption{From Narrow AI (fixed-purpose) to AGI.}
    \label{fig:avocado1}
\end{figure}

With the goal of providing a definition of GPAIS based on current research and articulating its differences with respect to traditional fixed-purpose AI and futuristic AGI, this section describes the main properties and functionalities one would expect for GPAIS, distinguishing different levels of ability and autonomy. Figure \ref{fig:avocado1} represents the transition from fixed-purpose AI to AGI.

Let $T$ be an AI/ML task, for which there is some data available and can be modelled in various ways depending on the expected outcome. For example, a classical ML task could consist of classifying images to distinguish among different types of objects (e.g. \emph{car} vs \emph{motorbike}). It is important to note that different ML tasks could be derived from the same source of data. A \textbf{fixed-purpose AI system} would consider a single task at a time, and would normally require having sufficient data to train a model. To solve a task, experts would typically follow the ML lifecycle, considering various strategies to learn effectively from the data (e.g. finding a suitable pipeline of preprocessing and learning algorithms, together with the tuning of their hyper-parameters). Having said that, the methodology proposed by an expert could be further applied to other tasks of different nature, as soon as they can be modelled within the same ML setting (e.g. as a single-output classification problem). For example, it is common to apply the same classification technique to various datasets with very minor modifications (e.g. some hyper-parameter tuning). Nevertheless, each task is tackled in a holistic way, training with enough data, and without any interaction/knowledge extracted from other tasks.

When shifting towards GPAIS, we simultaneously consider $N$ tasks ($T_1, \dots, T_N$). Thus, a GPAIS is expected to be able to solve one or more problems. While intuitively this means a more complex learning environment, dealing with a variety of tasks may allow us to exploit synergies among them (if they are sufficiently related). A classical example of this is multi-task learning \cite{Zhang2018}, which aims to improve the performance on multiple tasks by leveraging shared representations and knowledge transfer between them. 

Apart from dealing with more than one task at a time, to further characterise GPAIS let us  involve time in the definition. Let $t$ be the current point in time in which we are given a set of tasks ($T_1, \dots, T_N$). In GPAIS, a new set of $K$ tasks ($T_{N+1}, \dots, T_{N+K}$) may arise at a future time $t + \Delta t$, for which we do not necessarily expect to have much data to learn from. Within this setting, we differentiate between \textbf{closed-world GPAIS} and \textbf{open-world GPAIS:}
\begin{itemize}
    \item \textbf{Closed-world GPAIS}: In this type of GPAIS, we assume that we have data for a fixed number of tasks at a time $t$, and those will always be the only tasks to solve. It assumes that all the tasks that an AI system will encounter are predetermined and accounted for during the training phase. In a nutshell, we consider closed-world as the simplest form of GPAIS, which is capable of dealing with more than one task. However, these systems may lack the ability to generalise to novel tasks outside their training scope. Therefore, we would not expect a closed-world GPAIS to accomplish any task that it was not directly trained to do. If a new task arrived at the time $t + \Delta t$, the whole system would require to be retrained in order to perform well.
    
    \item \textbf{Open-world GPAIS}: This type of GPAIS refers to a scenario where the AI system operates in a more dynamic and evolving environment, encountering new tasks at a time $t + \Delta t$ and data that was not necessarily included at the time $t$ (namely, in its initial training set). Unlike closed-world GPAIS, open-world GPAIS acknowledges the presence of unknown and unforeseen tasks that may arise over time. This requires the AI system to possess a degree of generalisation, flexibility, and the ability to learn from limited or scarce data. To do so, a GPAIS would typically have to exploit what it learned previously at time $t$ to adapt faster to new tasks. \revision{Similar to the way human perception is inherently limited about what they know of the world, GPAIS operating in open-world scenarios also possess a substantial but constrained understanding of their context. This knowledge of the unknown could be treated as ``meta-information'' or knowledge about the environment, which could be leveraged to develop adaptation strategies to proactively confront the emergence of new tasks.}

\end{itemize}

Therefore, open-world GPAIS provides a higher degree of generality compared to closed-world. Looking at the current progress in the literature, we identify systems with various properties in terms of what they can or cannot do, or how they respond to the end user during certain tasks. In what follows, we make a few remarks with respect to data availability in both closed and open worlds, and the performance on new tasks:

\begin{itemize}
    \item \textit{Data availability}: Although it may not always be the case, in a closed-world GPAIS system we would normally expect sufficient data to train models on all the tasks. That does not imply that data augmentation and other pre-processing techniques may not be needed. However, in an open-world GPAIS, data scarcity may become the norm for new tasks that arise at time $t + \Delta t$. This is not always the case as, for example, many AutoML techniques usually assume enough data at time $t + \Delta t$, where an ML model is actually created. Further discussion on this is provided in Section \ref{sec:gpais-taxo}.

    \item \textit{Performance on new tasks}: When asked to perform new tasks, an open-world GPAIS may display difficulties while trying to solve them, but may not be aware of having such a difficulty. For example, in LLMs, the system may hallucinate and output untrue statements with high confidence \cite{Alkaissi2023}, or in computer vision, a generative model may produce images resembling objects that contain unrealistic elements \cite{Yu2021}. While hallucination in AI is not yet fully understood, several factors play an important role, including lack of data or biased training data \cite{Ji2023}. Conversely, we might also encounter a situation in which the system would successfully (yet serendipitously) perform tasks for which it was not intentionally and specifically trained (the so-called \emph{emergent} abilities) \cite{wei2022emergent}. \revision{These two issues raise concerns about the trust we may place on this kind of general-purpose system. Section \ref{sec:discussion} discusses in more depth the implications of these yet to be solved issues of open-world GPAIS.}
 
\end{itemize}

\begin{figure}[htp]
    \centering
    \includegraphics[width=\linewidth]{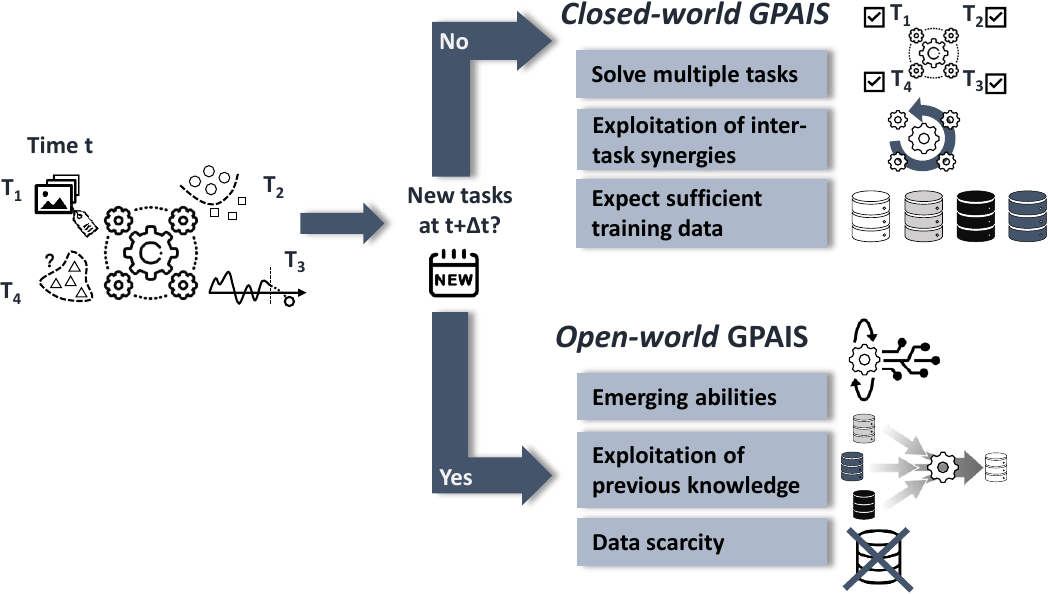}
    \caption{Closed-world vs. Open-world GPAIS: Some of their potential properties and characteristics.}
    \label{fig:gpais}
\end{figure}

Figure \ref{fig:gpais} shows a graphical representation of the differences between closed-world and open-world GPAIS, focusing on the time difference in which they work on, and noting some of the potential properties and characteristics of these systems. Note that some of those characteristics are not compulsory to qualify as open-world or closed-world GPAIS. For example, not all methods will have emerging abilities, or would be able to work well under data scarcity, but they may still be classed as open-world. The key distinguishing feature between closed-world and open-world lies in the expectation of new tasks at time $t + \Delta t$.

\begin{figure}[htp]
    \centering
    \includegraphics[width=\textwidth]{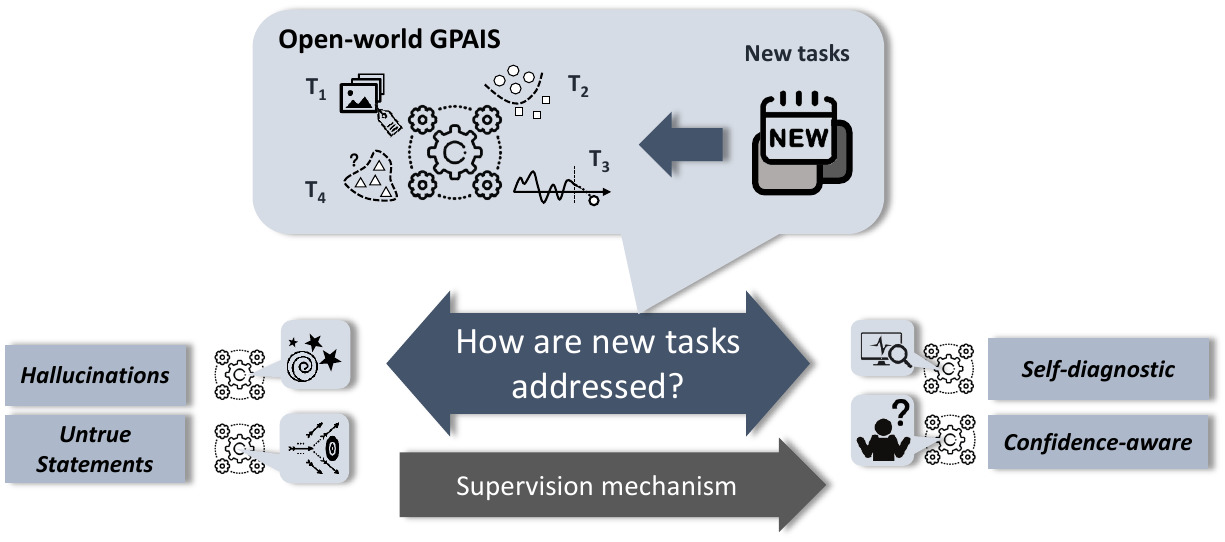}
    \caption{Open-world GPAIS: challenges and potential solutions to become advanced models.}
    \label{fig:levelsgpai}
\end{figure}

For a GPAIS to become advanced, we would expect them to respond more positively to new tasks, since they either have the information to perform the task in question or recognise that they do not know how to perform it. Advanced GPAIS would be expected to self-diagnose themselves, be aware of the confidence in their responses and ask (or look) for feedback and improve themselves within an active learning framework \cite{Budd2021} (See Section \ref{subsec:enrich}). To do so, a supervision mechanism could be introduced, so that, it learns to distinguish between what is correct and what is not. To date, none of the existing GPAIS provide such a level of autonomy. Figure \ref{fig:levelsgpai} illustrates the challenges posed by new tasks in an open-world setting, and how a supervision mechanism may help boost their confidence and capabilities. 

The final step in the evolution of these systems is to know under what conditions an advanced GPAIS would become an AGI system. An AGI system is characterised by being able to perform any task efficiently and as intelligently as a human being. It is precisely this quality of being human that advanced GPAIS need to become AGI systems since they must be able to do more complex reasoning, exploit causality, have a curiosity to explore new tasks, or have a conscience of themselves, among other abilities.

The properties and expected functionalities we have described before for GPAIS will be used to present a formal definition for GPAIS that considers different degrees of autonomy and ability:

\begin{definition}[GPAIS] 
A General Purpose Artificial Intelligence System (GPAIS) refers to an advanced AI system capable of effectively performing a range of distinct tasks. Its degree of autonomy and ability is determined by several key characteristics, including the capacity to adapt or perform well on new tasks that arise at a future time, the demonstration of competence in domains for which it was not intentionally and specifically trained, the ability to learn from limited data, and the proactive acknowledgment of its own limitations in order to enhance its performance.
\end{definition}

This definition fills the gap left by the definitions presented in the previous section, distinguishing between various degrees and properties of GPAIS. This definition may serve as a guide to transforming an existing AI system into a GPAIS.

\section{General Purpose Artificial Intelligence Systems: A Multidimensional Taxonomy} \label{sec:gpais-taxo}

The above definitions allowed us to distinguish between various degrees of GPAI. The aim of this section is to present a taxonomy of approaches to realise GPAIS of any kind, describing some of the key research trends and problems in which GPAIS are being developed. Although we will highlight how existing solutions fit within our definition and categorisation of GPAIS, it is important to early note that there may be some degree of overlapping between categories, and certain strategies can facilitate the transition of methods across categories (e.g. adding some elements to evolve a typically closed-world AI system into an open-world one). It should also be stressed that in this work, our primary target is on the ML aspect of AI, which encompasses other research areas such as optimisation or simulation. 

\subsection{Breaking it down onto a multidimensional taxonomy}
\label{subsec:taxonomy}

With the aim of making AI more self-sufficient and capable of learning without human intervention, researchers in the AI community have followed many strategies to provide generalisation abilities. Broadly speaking, we observe two distinct approaches in the literature: \textbf{AI-powered AI} vs \textbf{single-model} approaches. 

\begin{itemize}
\item \textbf{AI-powered AI}:
We can consider an additional layer of abstraction that would use an AI algorithm to either design or enrich another AI algorithm. As an example, the hybridisation of ML techniques and optimisation algorithms, by interacting with each other or themselves \cite{Song2019}, has recurrently been exploited to foster generalisation (among other objectives). In essence, we are boosting the performance and robustness of an underlying AI model via another AI technique. Throughout the following subsections, we will discuss the different ways in which AI may power other AI systems, categorising AI-powered AI as per their objective: designing an AI algorithm or helping/enriching an AI algorithm to learn/perform better.

\item \textbf{Single-model}: On the other hand, not all the advances in GPAIS would always need to use an extra AI model to help generalise. Instead, their generalisation abilities come from learning from various tasks and/or vast amounts of data. We identify two relevant research areas that at their original definition would typically use a single AI model, namely, multi-task learning and foundation models. It is important to clarify that while we categorise these techniques differently from the AI-powered AI ones, that does not mean they offer fewer abilities, nor could they be combined with the above in many cases. 
\end{itemize}

\begin{figure}[!h]
\includegraphics[width=\textwidth]{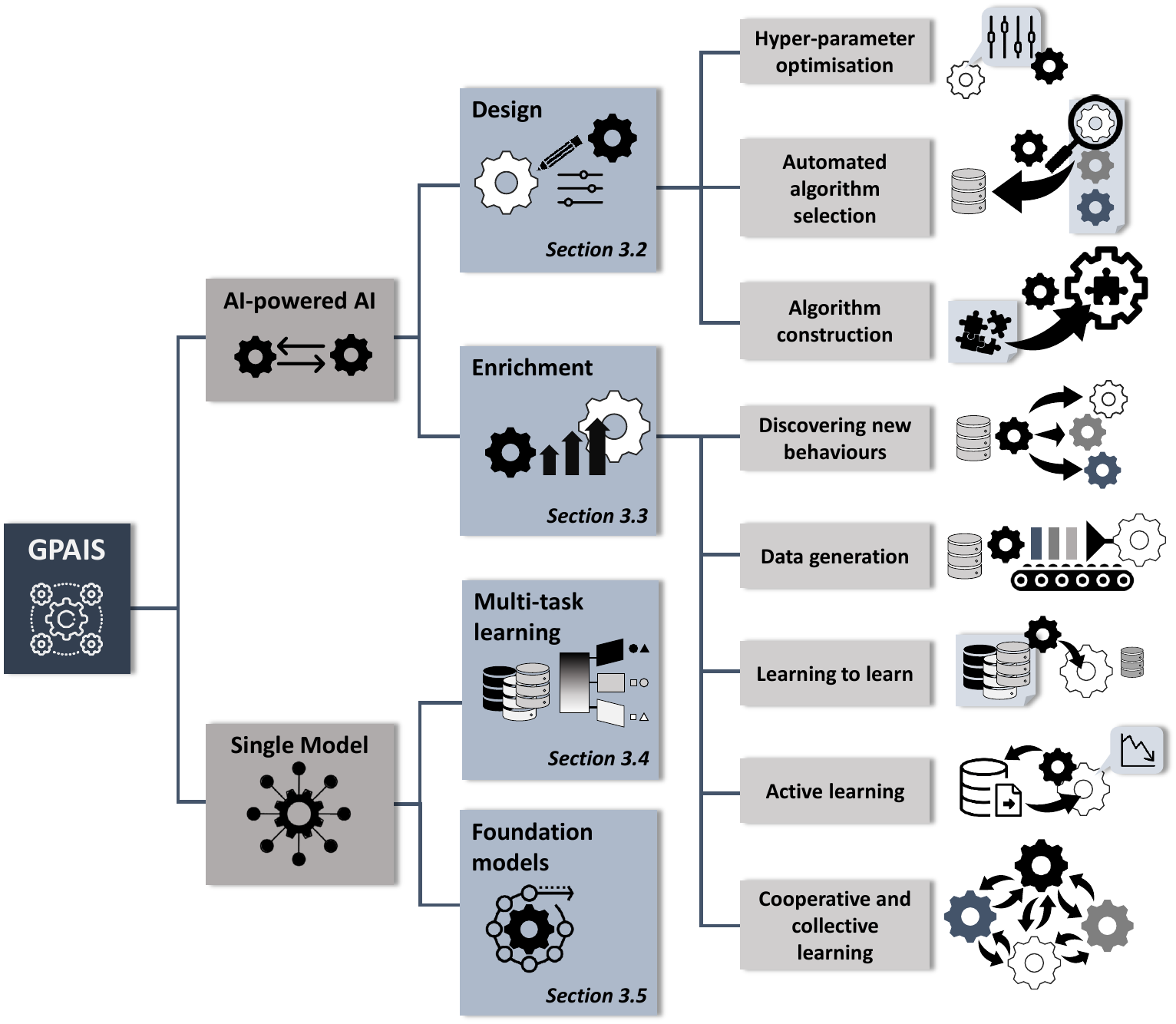}
    \caption{Taxonomy of approaches for GPAIS.} 
    \label{fig:taxonomygpais}
\end{figure}

Figure \ref{fig:taxonomygpais} shows the proposed multidimensional taxonomy for different categories of GPAIS, approaches and research areas that we discuss in the following subsections. Sections \ref{subsec:design} and \ref{subsec:enrich} summarise some of the ways in which AI can be used to make another AI system more autonomous and general to design AI algorithms and to enrich other AI models, respectively. Sections \ref{subsec:multi-task} and \ref{subsec:foundation} cover multi-task learning and foundation models, respectively, as prevalent examples of single AI models that generalise to multiple tasks, but do not necessarily involve an additional AI layer. 

\subsection{AI to design AI}
\label{subsec:design}

To solve a task with an AI system, experts are typically confronted with various design decisions that range from customising the hyper-parameters of a well-known algorithm, selecting an appropriate method and its hyper-parameters, to designing the components of -- or even the entire -- AI algorithm. It is common for the stages of the ML Lifecycle to become somewhat repetitive, so experts often base their decision on their own experience with previous problems. AI algorithms can be used to help with all of those design decisions, extracting general knowledge about how to implement an ML process easier and faster within a set of ML tasks.

\begin{itemize}
\item \textbf{Hyper-parameter optimisation}: Determining the best hyper-parameters for a given task is of extreme importance to achieve good results. While traditional hyper-parameter tuning strategies have frequently been used to tackle a specific dataset and that would not constitute general-purpose, they can also be used to tune hyper-parameters for related datasets \cite{Barros2014}. To do this, metaheuristics and Bayesian optimisation are well-known examples of commonly used techniques for black-box function optimisation \cite{Snoek2012}. Among others, general frameworks such as Optuna \cite{Akiba2019} or Hyperopt \cite{Bergstra2015} greatly facilitate the search for adequate hyper-parameters. 

  In GPAI, we could find the best set of hyper-parameters for a number of tasks, or even determine which of them work best for different tasks, and use that knowledge for when a new task arises. Thus, this can be seen as exploiting previous knowledge and generalising beyond the original tasks over which hyper-parameter tuning was done. Nevertheless, if the new task has very little data, adapting/fine-tuning those hyper-parameters could become very challenging.
    
    \item \textbf{Automated algorithm selection}: As there are a wide variety of AI algorithms, selecting the best model(s) is also critical in solving a task. In the world of GPAIS, the previous hyper-parameters optimisation is often coupled together with the automated selection of an AI algorithm (also known as the combined algorithm selection and hyper-parameter problem \cite{Thornton2013} in ML). Depending on the kind of AI we aim to further automate, e.g. ML or optimisation, we may find different terminologies to refer to this kind of strategy. In ML, they are typically referred to as \textbf{AutoML} \cite{Hutter2019}, while in optimisation, they would usually be known as hyper-heuristics \cite{Burke2013}. In ML, it is common for the AutoML to optimise an entire pipeline, involving the automated selection of the best combination of a pre-processing technique (e.g. one-hot encoding, dimensionality reduction, or feature selection, among others), as well as an ML algorithm  \cite{Feurer2020}. \revision{In the area of reinforcement learning, the idea of AutoML may also be considered to automate the reinforcement learning pipeline with some specific challenges (e.g. non-stationarity, environment and algorithm design, or the generation of complex and diverse behaviours) \cite{ParRaj2022}.}

    Many authors talk about AutoML \cite{HE2021} when the aim is to find the best pipeline for a deep learning approaching, covering, feature engineering, hyper-parameter optimisation and neural architecture search \cite{Elsken2019, schrodi2023construction}. In this field, metaheuristics have made a good contribution through the use of evolutionary algorithms, aiming to both reduce the complexity of the deep learning model and obtain better results \cite{Martinez2021, Zhan2022}. \revision{As an alternative to finding the best pipeline, \cite{hollmann2023tabpfn} has recently shown that a transformer \cite{Vaswani2017} can be used to perform classification without designing explicitly a machine learning pipeline.}

    While the common objective of these techniques is to take the human out of learning/optimisation process \cite{Quanming2018}, different methods may be only capable of dealing with closed-world GPAI, and others may work in open-world scenarios. Some may use previous knowledge available at time $t$ (e.g. those based on meta-learning), whereas others may focus on adaptation to the new task, aiming to find the best algorithm at time $t + \Delta t$. As happened before for hyper-parameter tuning, when tackling open-world tasks, current automated algorithm selection techniques would normally work under the assumption that sufficient data is still provided to create a model for the new task. 
    \revision{One could argue, however, that a meta-learning approach could be viable to find the best algorithm and hyper-parameters within a zero-shot learning approach, which assumes that some meta-information about the new task is available and that it can be related to previously known tasks. Similar to the way auto-sklearn works \cite{Feurer2020}, this could be in the form of meta-features for the new dataset which describes its main properties.}

    \item \textbf{Algorithm construction}: The previous two approaches are focused on the configuration of existing algorithms. However, in addition to these approaches, various AI strategies have been proposed to go beyond algorithm configuration and actually construct new algorithms from scratch or determine the necessary low-level components for a given task.

    AutoML-zero \cite{real2020automl} and Hyper-heuristics \cite{Wenjie2022} are two examples of AI-based strategies to generate new algorithms for ML and optimisation, respectively. Neuroevolution \cite{Stanley201924} is also an alternative to designing entire deep learning algorithms from scratch. AutoML-zero explores the space of possible algorithms and their configurations using search algorithms. This strategy enables the creation of novel algorithms tailored to specific tasks or problem domains. Current developments are focused on designing algorithms for a set of tasks within a specific domain, which would class as closed-world within the proposed definition. However, there is potential in the future to become open-world, by evolving the generated algorithms in new tasks, or by formulating new objective functions in existing algorithmic construction networks that seek generalisation over unseen tasks.

\end{itemize}

\subsection{AI to enrich AI}
\label{subsec:enrich}

Instead of focusing solely on the design of AI algorithms, AI is often coupled with other AI techniques to enrich or help with its learning process. By harnessing the power of various AI techniques and their collaborative potential, we can empower AI systems to become more robust and capable intelligent systems. In this context, this section identifies some of the most relevant areas in which AI may help make another AI to be more general-purpose and adaptable.
\begin{itemize}
    \item \textbf{Discovering new behaviours}: One of the challenges for an AI is to dynamically adapt to changes in the environment, such as shifts in the underlying data distribution. \textbf{Continual learning} \cite{Parisi2019, Delange2022} is an ML paradigm that involves learning a model able to solve several tasks from a continuous stream of data, without forgetting knowledge obtained from the preceding tasks. Data related to the old tasks are not available during the training of the new tasks, forcing the adaptability of the system and the preservation and retrieval of valuable knowledge over time. By leveraging other AI techniques, we can explore innovative approaches to discovering new behaviours. Research on data stream mining has largely addressed concept drift detection and adaptation methods \cite{Gama2014} to help identify instances where the data distribution has changed significantly, and to efficiently adapt the knowledge captured by the model to such changes. Thus, it can be viewed as an indicator of new behaviours or patterns in the data. Other ways to facilitate discovering and adapting, avoiding catastrophic forgetting, involve reinforcement learning and meta-learning \cite{Khurram2019}. For instance, reinforcement learning \cite{Xu2018} could help explore and learn from interactions with the environment, allowing for the emergence of novel and adaptive behaviours that could help the agent performs better in varying tasks and/or new environments. \revision{Another area in optimisation research with ample potential to discover emerging novel behaviours in data-based models is quality-diversity optimization \cite{chatzilygeroudis2021quality,cully2017quality}. Techniques belonging to this research area permit to efficiently explore complex search spaces by simultaneously accounting for the quality (fitness) and diversity (similarity) of their improvised solutions. When applied to the optimization of the parameters of a given model (as in Deep Learning), quality-diversity optimization can yield diversified GPAIS that, when combined together, may effectively cope with unseen modeling tasks in few- or even zero-shot open-world settings, especially when such new tasks are not radically different from the ones providing the training data for the GPAIS}. 
    \revision{These sorts of GPAIS techniques may belong to both the closed-world and open-world settings depending on their abilities. On the one hand, when the tasks themselves do not really change, but the distribution of the data does (i.e. a new concept emerges), it would be classified as closed-world. On the other hand, they could also belong to the open-world setting when they are capable of dealing with entirely new tasks \cite{Cuong2018}, as in robotic tasks in which the action space changes over time.}. 
    

    \item \textbf{Data generation}: One of the cornerstones in AI to generalise well is the lack of data that could resemble what might happen at time $t + \Delta t$. With the emergence of \revision{GenAI, we find} generative models \cite{Stokel-Walker2023} that are capable of creating data that follows the distribution of a given training dataset\revision{, achieving unprecedented scales and levels of fidelity over complex data modalities}.  \revision{This makes it possible to simulate and generate potential data samples (and environments in a reinforcement learning context)}. This opens up new avenues for improving AI's ability to generalise well by augmenting the existing dataset and expanding the range of scenarios the model can learn from. Additionally and linked to the previous challenge, creating synthetic data can facilitate the discovery of new patterns, behaviours, or anomalies that may not have been explicitly present in the original training, further enhancing the AI's capacity to generalise effectively. \revision{In the context of reinforcement learning, Paired Open-Ended Trailblazer (POET) \cite{wang2019paired} is a prime example of environment generation to improve the generality of a model in potential open-world scenarios.}  In \cite{Clune2020}, the author highlights that learning how to automatically generate effective training data is one of the most important pillars to realise generalpurpose intelligence, and it is currently the most underdeveloped area. More about generative models will be discussed in Section \ref{sec:generativemodels}.


    \item \textbf{Learning to learn}: Having many (sometimes related) tasks to deal with and the lack of data have motivated the research on AI systems that learn how to learn. As mentioned before, meta-learning \cite{Lemke2015} finds applications in the design of AI (e.g. hyper-parameter optimisation and AutoML), but it is also applicable to other scenarios in which it is designed to help other AI models learn. The ways in which meta-learning can help other AI methods may range from learning an efficient optimisation algorithm for the sake of faster convergence and better performance \cite{Li2017}, which would not class a general purpose system, to learning how to transfer knowledge effectively from a set of tasks to a new task \cite{Wang2017}. In GPAIS, \textbf{few-shot learning} \cite{Wang2020} is a prime example of beneficiaries of meta-learning approaches. The challenge in the few-shot regime is to learn effectively (without overfitting) from an extremely limited number of labelled examples (or even none, known as \emph{zero-shot} learning \cite{Xian2019}). Although few-shot learning can be tackled without meta-learning \cite{Koch2015, Vinyals2016}, many methods do use it to mimic human learning. These methods extract meta-knowledge from a collection of few-shot learning tasks, and then transfer this meta-knowledge to unseen few-shot learning tasks comprising novel categories. In this way, there are two distinct AI models at play: the meta-learning model and the base model. The meta-learning model is responsible for acquiring meta-knowledge from multiple tasks, while the base model leverages that meta-knowledge to perform few-shot learning on new tasks. Thus, the majority of the methods that follow this learning-to-learn paradigm may naturally class as open-world.

    \item \textbf{Active learning}: Another of the weaknesses of AI is to realise when they are wrong and seek human intervention. Active learning \cite{Settles2012} has traditionally been used to reduce annotation effort by intelligently selecting the most informative instances for labelling. Next, the algorithm being trained asks an \textit{oracle} (which is usually human assistance) to label new data that may help to improve its performance, with the goal of maximising the performance of the algorithm with the least amount of data. This idea could potentially be very useful in GPAIS when the demand for new annotated data is autonomously decided by the model itself \cite{Fang2017}. Thus, it can also facilitate adaptability by enabling an AI system to actively seek guidance from humans or oracles in unfamiliar or uncertain situations. It allows the AI system to acquire new knowledge and fine-tune its performance to different tasks, enhancing its generalisation capabilities. This has the potential to develop advanced open-world GPAIS that are closer to AGI. \revision{To realise an effective active learning approach, GPAIS should have sufficient reasoning tools to identify the need for guidance. In this context, knowledge graphs and semantic databases could provide the understanding needed to query for the right data labelling \cite{khan2023knowledge}.}
        
    \item \textbf{Cooperative and collective learning}: To mitigate the weaknesses of different AI systems alone, we could get them to help each other to constitute a larger, general purpose system. Collaborative AI refers to the concept of multiple AI agents working together, often in a coordinated manner, to achieve common goals or solve complex problems. While this is not a new research area \cite{Panait2005}, in the context of GPAIS, collaborative AI agents/models can be employed to enhance the capabilities and performance of the system. These agents could specialise in different tasks or domains and collaborate with each other to provide a more comprehensive and versatile solution. For example, within a GPAIS, different AI models could focus on tasks such as natural language understanding, image recognition, recommendation systems, or decision-making, and they can exchange information or cooperate to address complex queries or problems.

    Related to the idea of collaboration between AI systems, we may find \textbf{Federated Learning} \cite{Silva2023}, as a decentralised ML approach that enables multiple devices or nodes to collectively train a shared model without sharing their raw data. In the field of robotics, the concept of \textbf{Swarm Intelligence} \cite{Dorigo2021} has gained attention for its potential in achieving general intelligence \cite{Kwa2023}. Swarm Intelligence involves the emulation of collective behaviours observed in social insects, where individual agents interact locally and make decisions based on simple rules or local information. The majority of existing federated learning or swarm intelligence methods would not class as GPAIS as they are usually developed to perform a single task. Nevertheless, the ability of these distributed AI models is not limited to fixed-purposed models, and GPAIS with other attributes such as being capable of dealing with multiple tasks and/or continual learning \cite{Zweig2017} could be built using a decentralised approach. Therefore, the categorisation of a cooperative and collective learning approach as closed-world or open-world depends on the features of the underlying model.

\end{itemize}

\subsection{Multi-task learning}
\label{subsec:multi-task}

Multi-task learning is an ML technique that allows a model to be trained simultaneously on different tasks, exploiting the synergies between them. Consequently, knowledge and representation are shared across all tasks with the goal of maximising model performance \cite{Zhang2018}. When learning from different domains of data, a more robust behaviour for new data is expected \cite{Ghifary2015, Sosnina2023}, even against adversary attacks \cite{Chen2022}. 

However, multi-task models may face a higher risk of error and lower efficiency in tackling all tasks compared to models trained individually for each task. This problem has been studied through different approaches, as shown in \cite{Zhang2018, Zhang2022}. Although there have been certain techniques that have proven to be good options such as metaheuristics and Reinforcement Learning \cite{Martinez2022}, currently deep learning models like Gato \cite{reed2022} and the more recent Meta-transformer \cite{zhang2023metatransformer} have shown to be the most successful approaches for multi-task learning, especially with different data modalities and diverse learning problems.

Traditional multi-task learning algorithms would typically fall under the closed-world GPAIS category because a multi-task system is trained in a stationary environment, i.e. it is trained to solve a set of tasks at a time $t$. However, it is not usually conceived to tackle new tasks. If a new task emerged, data related to that task would have to be added. This would normally imply retraining the model as a whole, although other strategies are possible. These strategies to enable multi-task learning in an open-world setting would require a good amount of data for the new task in order to perform well. In contradistinction to retraining the model, we could follow a pre-training plus fine-tuning (as in \cite{Chebotar2021}) to make a multi-task learning model perform well in the open world, giving rise to the next category discussed in the next section.

\subsection{Foundation models}
\label{subsec:foundation}

As discussed previously, the underlying idea of foundation models is that a single model is trained on a very large amount of data (typically for multiple tasks), and in a later stage (time $t + \Delta t$), such a model can be fine-tuned to tackle new tasks. As described in \cite{Bommasani2021}, these models are based on \textbf{deep learning} and \textbf{transfer learning}. These methods have become very popular in LLMs, demonstrating strong general purpose abilities to generate human-like text, which will be the main topic of the next section.

It is important to clarify that both meta-learning and transfer learning involve leveraging knowledge from previous tasks. However, meta-learning focuses on developing learning algorithms or models that can learn how to learn, adapt, and generalise, while transfer learning aims to transfer specific knowledge or representations from a source task to improve performance on a target task. Although the purpose does not directly relate to learning to learn, the training and adaptation framework of foundation models facilitates the separation of concerns. For example, in the context of natural language processing, the training process of an LLM like ChatGPT may involve exposure to a vast range of text sources, such as books, articles, and websites. During this training process, the model learns various aspects of language, including grammar, syntax, semantics, and common word usage. This meta-knowledge acquired during training allows the model to understand and generate coherent and contextually appropriate text.

\revision{One of the keys to successfully training foundation models lies in how to use vast amounts of unlabelled data effectively. Semi-supervised learning -- and more prominently, self-supervision-- have been key research areas to deliver high-quality foundation models. Self-supervised learning \cite{balestriero2023cookbook} allows us to reduce the necessity for labels by creating its own labelled data and solving auxiliary (pretext) tasks to extract knowledge therefrom. For example, in LLMs, it is common to mask a word out in the text and predict the surrounding words. This permits us to model the relationships among consecutive words without the need for external labels. These relationships can later be used for downstream tasks, such as text generation or translation to other languages.}

Once the foundation model is trained, it can be adapted to perform specific tasks or domains. For instance, it can be fine-tuned on a dataset of movie reviews to learn how to classify sentiments (e.g., positive or negative) expressed in film reviews. The adaptation process involves providing task-specific data and adjusting the model's parameters to make it more specialised and accurate for the particular task at hand, retaining as much previous knowledge as was found to be valuable for the specific task/domain under the target.

Foundation models are inherently open-world, but they may have some limitations. For example, they require enough quality data for the adaptation phase to be successful. When the amount of data is low, meta-learning comes into play. We could leverage a foundation model to quickly adapt to new tasks with limited data. Even with the addition of meta-learning to tackle new tasks with very small amounts of data, we would not class foundation models as advanced GPAIS. For them to become advanced models, they should be hybridised with other AI models, such as active learning.


\section{A closer look into Generative AI: a kind of Foundation Models in GPAIS}
\label{sec:generativemodels}

As mentioned in the introduction, modern GenAI models like ChatGPT are the first AI models that the general public has begun to recognise as GPAIS. Therefore, due to their relevance, and to align them with the terms and concepts presented in the taxonomy of Section \ref{sec:gpais-taxo}, we now provide a more in-depth vision of these approaches.

As commented above, GenAI models \cite{Stokel-Walker2023} are an outstanding example of how AI may help enrich other AI systems. \textbf{\textit{Generative modelling}} or \textbf{\textit{GenAI}} consists of a set of algorithms designed to learn the distribution of a dataset, so that, its underlying patterns can be characterised, and samples that resemble the original data can be generated \cite{Oussidi2018}. Queries for new data instances can be performed unconditionally or conditioned on a query to produce samples with specific traits (e.g. class belongingness, the induction of high-level properties on the output, or a descriptive prompt, among many other possibilities).

In the literature, one of the earliest generative models that we may encounter are generative adversarial networks (GANs) \cite{Goodfellow2014}, which have showcased a great ability to generate highly specific data, including realistic images \cite{Yu2021} and medical images \cite{McAlpine2022}. However, by design GANs are only trained to improve a specific type of results, because the generation process is evaluated/trained with a discriminator for a very specific task \cite{Toutouh23}. As such, they can not be considered enablers for general purpose intelligence, but they could help improve AI systems generating additional data to learn as described in Section \ref{sec:gpais-taxo}.

Recently, various AI models have been proposed that, differently from GANs, are able to generate a wide variety of images, ranging from paintings to photo-realistic images, based on textual descriptions. At their core, these text-to-image models are diffusion probabilistic models or diffusion models (DMs) that capture a higher-level semantic meaning of a group of images \cite{Rombach2022}, so they are considered generative models. LLMs are also considered generative models because they are trained using many texts for which they learn the probability distribution over its vocabulary. Once trained, LLMs are able to generate content similar to original texts, so they seem to have been written by a human. In essence, both DMs and LLMs are \textbf{Foundations Models} as they are trained on large quantities of unlabelled text containing up to trillions of tokens (for LLMs) or millions of pictures (for DMs), using self-supervised or semi-supervised learning. These models, in contrast to GANs, are not trained for a very specific task, and they are able to solve many problems \cite{Stokel-Walker2023}, so they can be considered GPAIS. \revision{It is important to highlight that DMs may not necessarily be GPAIS if their focus is merely on generating images, and if that is their only task. LLMs, however, can easily be used for multiple tasks (e.g. translation, summarising, etc).}

Some of the most relevant DMs include DALL-E \revision{\cite{DALLPretrainingMitigationsa,betker2023}}, Stable Diffusion \revision{models (as 2 \cite{Rombach2022} or XL \cite{podell2023}}), and the commercial tool Midjourney. These models have demonstrated their capability to create professional images solely from textual descriptions, making them advanced tools for image generation. Their creations are of such high quality that experts might even consider using them to win an art competition \footnote{\url{https://www.nytimes.com/2022/09/02/technology/ai-artificial-intelligence-artists.html} (Last access: 2023/10/28)}.

Currently, the more relevant generative models in languages are Bidirectional Encoder Representations from Transformers (BERT) \cite{Devlin2019} and Generative Pre-trained Transformer (GPT). In terms of architecture, BERT is a bidirectional model that can consider both the preceding and succeeding text when generating output. It consists of encoder and decoder components. In contrast, GPT is a unidirectional model that generates subsequent text based on the context and preceding text. Although BERT can be used for question-answering tasks, particularly for specific questions, it was not primarily designed for generating free-form text with coherent and contextually relevant content, unlike GPT models. Therefore, we will focus on these latest models. GPT models are able to generate human-like text based on prompts given to them. They can participate in conversations, answer questions, provide explanations, and assist with various language-related tasks. GPT has the ability to adjust its output based on the context of the conversation, enabling it to provide more accurate answers. The following LLMs (in all their versions) share similarities with GPT, but they differ in terms of the underlying models used for text generation and the training processes employed.

The most relevant LLMs are ChatGPT \cite{Vandis2023}, GPT-4 \cite{Openai2023} (both proposed by OpenAI in collaboration with Microsoft), LLAMA 1 \cite{Touvron2023v1} (proposed by Meta) which has produced a list of improved models with additional fine-tuning like Vicuna \cite{Chiang2023,Bubeck2023}), LLAMA 2 \cite{Touvron2023v2} or Google AI Bard \cite{Patrizio2023}. These models achieve unprecedented performance due to a two-phase approach: a training process involving a large amount of textual information (obtained from the web and other resources, some of which are not specified), and a reinforcement learning stage involving human input. This combination enables AI models to generate text that closely resembles human-authored writing. LLAMA 1 lacks the aforementioned reinforcement learning phase, resulting in less natural output. To address this limitation, tuned models have been developed, such as Vicuna. However, in the case of Vicuna, reinforcement learning was conducted by a GPT model instead of humans due to resource constraints. The use of an LLM as ChatGPT to evaluate the output of other LLMs can be considered an interesting case of an AI model to enrich AI, although some authors do not recommend this evaluation methodology \cite{Gudibande2023}. \revision{LLAMA 2, however, has a version tuned for chat conversations, which has been coined as LLAMA-2 Chat \cite{Touvron2023v2}}.

Unfortunately, the actual GPT-4, ChatGPT or Bard models are not really available to the end user. Therefore, it is not possible to have a copy of the models to use them locally or to adapt them using fine-tuning. The only way to use them is through a remote interface that enables users to interact with the AI model hosted on a remote and private server. LLAMA 1 and their fine-tuned version are available for local use in various sizes, but they are not permitted for commercial purposes. Nevertheless, there is an increasing number of models that allow it, like the new LLAMA 2 \cite{Touvron2023v2} \revision{(for a limited number of active users)}, Falcon LLM \cite{Penedo2023}, or Dolly, from Databricks \cite{Conover2023}.

One of the most notable features of these generative models is their ability to synthesise information from various domains and their capacity to apply knowledge and skills in diverse contexts and disciplines. Some models, such as GPT-4, have demonstrated a certain level of intelligence \cite{Bubeck2023}. First, these models have shown a high level of proficiency in various domains such as literature, mathematics, programming law, and others. This enables their use not only for general tasks but also for highly specialised ones. However, these multi-domain models coexist with other specific models for some domains of special interest. As an example, Github CoPilot is an AI model specifically trained to be capable of generating source code, used as an advanced development tool that can improve the productivity of developers \cite{Dakhel2023}.

Another significant feature of these generative AI models is their multi-modality, which enables them to combine multiple modalities and generate outputs that are more diverse and nuanced. This implies the ability to combine output and relationships between different formats such as text, images, audio, and more. For example, a multi-modal AI model can generate both a description of an image and an image based on text input. Furthermore, if it is trained on both text and audio, it can generate speech based on text input or generate text from speech input. This feature indicates that these generative AI models are not limited to a single type of information, thereby enhancing their potential to solve complex tasks. \revision{The next section will provide more information about multimodality in GPAIS.}

As foundation models, generative models are part of the open-world category, typically requiring enough data and a fine-tuning step to adapt to new tasks. Due to their striking performance and assertiveness, they can easily be confused with human intelligence \cite{Sejnowski2023}. However, these models may \textit{hallucinate}, providing very confident responses that do not seem to be justified by its training data, hence they could be completely false.
The main issue with hallucinations is that the outputs provided by a model may sound plausible but are either factually incorrect or even unrelated to the given context, raising concerns about how to trust these models and ethical implications \cite{DiazRodriguez2023}. \revision{This problem can be even more severe in cases where generative models are used to produce training data to learn other models \cite{shrivastava2017}. These techniques are specially interesting to generate synthetic medical images  \cite{kazuhiro2018,guan2019,bissoto2021}, for privacy and anonymity reasons \cite{bissoto2021,thambawita2021a}. However, due to the severity of the medical errors, the danger of hallucinations may be greater, requiring specialist supervision to minimise it.}
One may also argue that hallucinations may be related to speculation and therefore similar to how humans reason in case of a lack of information. Additionally, there is a debate whether this kind of model shows emergent abilities \cite{wei2022emergent, schaeffer2023emergent, Bubeck2023}, meaning that they are capable of solving tasks that they were not explicitly trained for. Both hallucinations and emergent abilities are two current hot topics in LLMs and DMs (See Section \ref{sec:discussion}).

A very interesting and promising approach is the use of LLM-powered autonomous agent systems, in which LLMs could be used as an intelligent tool to divide a problem using task decomposition \cite{Wei2023}. They use several LLM agents to solve different partial problems or even decision-making tasks \cite{Yang2023}, like AutoGPT\footnote{\url{https://github.com/Significant-Gravitas/Auto-GPT} (Last access: 2023/10/28)} or HuggingGTP \cite{Shen2023} or MetaGPT \cite{Hong2023metagpt}, or in which the AI model is able to interactively ask more questions, like GTP-Engineer\footnote{\url{https://github.com/AntonOsika/gpt-engineer} (Last access: 2023/10/28)}. This cooperative approach, which aligns with the potential of \emph{cooperative and collective learning} discussed in Section \ref{subsec:enrich}, has already shown interesting results in complex domains like chemistry \cite{Bran2023}.

\revision{
\section{Multi-modality in GPAIS} 
\label{sec:multimodal}

A significant challenge for real-world systems is the integration of data coming from multiple sources, called \textit{data fusion} \cite{Hall97}. AI has extensively been used for this task \cite{Meng2020}. Traditionally, this fusion implies merging different data from the same kind of source, for instance, medical images resulting from several tests should be merged when related to the same organs \cite{hermessi2021a,james2015}. However, the variety of data available for a given problem is rapidly increasing, including images, text, audio, etc \cite{zhu2020}. This multi-modality of data types provides an opportunity to GPAIS, which can learn a more cohesive representation of a concept by having multiple ``views'', very much in the same way that humans do. For example, a human learns what a cat is (i.e. the concept) by seeing, hearing, touching and even smelling it. GPAIS that only use a single source of data may be losing an important part of the concept (e.g. LLMs would only know what a cat is from what it has ``read''). Conversely, having such a variety of data poses a challenge to learning effectively.

In recent years, there has been a lot of effort in using AI models capable of processing multimodal data \cite{Baltru2019,zhu2020}. Modern deep learning models have also been researched in the context of multi-modality \cite{Gao2020}. A good example of multi-modality in GPAIS is Gato \cite{reed2022}, a general purpose agent that may use text and images. Recent models, like graph deep learning models \cite{wu2021,zhou2020}, can represent more structured data, enabling AI systems to tackle new prospects and applications \cite{Zhang2022b}.

In prompt-to-image GenAI models, there is always a certain multi-modality, because they receive some text as input and it must create an image. Consequently, two distinct media are involved. For LLMs, although they achieve great success only by working with text, the ability to process images could improve their functionality and, in particular, their abilities to communicate with the user (e.g. answering questions like describing a picture). Nowadays, this functionality has been recently added to ChatGPT, and is expected to improve the potential usages of this technology (e.g. to help people with vision impairment \cite{Felix2018,Konig2022} or even improve the training of other generative models, like DALL-E 3 \cite{betker2023}). However, the real challenge with multi-modal models is the integration of information coming from different sources about the same concept in a cohesive format, in a way that allows an AI model to have a more general view of the target concept. For instance, joining visual recognition (recognising an object) and conversational commands (processing it) can be very useful to improve the interaction with people \cite{gongMultiModalGPTVisionLanguage2023} and even in robotic environments \cite{Chen2022b}.

Multi-modality as a way to improve AI models is the approach for Google's new advancement in AI, spearheaded by their proposal called \textit{Generalised Multimodal Intelligence Network Interface model} (Gemini AI) \footnote{\url{https://www.gemini-ai.org/} (Last access: 2023/10/28)}. Gemini AI has been designed to be versatile, to be able to tackle diverse tasks and to learn from a myriad of domains without the current constraints. To this end, Gemini embodies an interconnected network composed of individual modular ML models, which are trained on specific tasks. The different modules will produce varied outputs, and the Gemini AI encoders will transform these diverse data forms into a cohesive and common format. Then, decoders will produce outputs in diverse modalities, contingent on the received encoded inputs and the task on focus. It is expected to greatly improve current multimodal AI models.

Gemini AI was not available at the time of writing this work. However, it is clear that Google has decided to focus its strategy on incorporating improved multi-modality in their AI models as a way to outperform current state-of-the-art GenAI models. This movement suggests that for renowned experts (it is the collaboration work of well-known AI labs, such as DeepMind and Google Brain) this could be a strategic field that could significantly expand the realm of possibilities for future AI models.}

\section{A Discussion on the Prospects,  Implications and Regulation and Governance of GPAIS} 
\label{sec:discussion}

The previous sections have aimed to provide a definition for GPAIS considering various degrees of autonomy (Section \ref{sec:gpais-defs}) and a taxonomy of methods to build these systems (Section \ref{sec:gpais-taxo}). In this section, we briefly discuss on the current state of GPAIS prospects and limitations, the implications of it in our society, and the need for regulation and governance.

\subsection{Current status and prospects} 
\label{subsec:prospects}
As discussed before, LLMs are currently the most prominent GPAIS we may find that display some degree of general intelligence. While they are certainly the most outstanding ones, LLMs are not the sole means of generalisation that have been extensively researched in the literature, which could significantly contribute to the development of a GPAIS. Our taxonomy offers a comprehensive perspective, encompassing various approaches and methods employed in building these systems. However, the proposed taxonomy is not intended to establish a sharp separation across methods, and we may find cases in which some AI systems may belong to more than one category at once, which could be desirable, as we will discuss next. While many considerations are being made in different research niches, we hope that our taxonomy facilitates a more holistic and global approach to GPAIS development.

To advance in the field of GPAIS, it is crucial to explore open-world approaches that enable systems to operate in dynamic and unfamiliar environments, where new tasks are dealt with limited data, exploiting as much as possible previous knowledge. In doing so, we advocate for considering not only very successful approaches such as foundation models, but also hybridising them with some of the other strategies discussed in this article. For example, foundation models such as ChatGPT have already rendered exceptional performance in complex tasks, which could be enhanced if other approaches such as \emph{AI to enrich AI} methods are used in conjunction with them. From the approaches discussed in Section \ref{sec:gpais-taxo}, we highlight \emph{continual learning} (i.e. models capable of discovering new behaviours in the data), learning to learn, and active learning as some of the most prominent approaches that could be used in combination with foundation models. 

By incorporating these approaches into GPAIS, we can enhance their adaptability, generalisation capabilities, and learning efficiency in the presence of new learning tasks:
\begin{itemize}
    \item Continual learning can enable GPAIS to keep up with varying data distributions, while meta-learning can endow them with the ability to quickly adapt to new tasks and domains. 

    \item Active learning and reinforcement learning empower GPAIS to actively seek information and learn from their interactions with the environment. Proactive measures such as acknowledging limitations and asking for help, or finding ways to look for additional sources of information, would help GPAIS to become more autonomous. 
\end{itemize}

Together, these techniques pave the way for more advanced and versatile GPAIS that can address complex real-world challenges effectively. For current GPAIS to evolve beyond their limitations, we must explore their potential for developing emergent abilities and addressing inherent challenges. These two aspects are calling for further research with the aim of understanding whether emergent abilities are real and explaining why they occur, together with means to identify and control hallucinations. 

\subsection{Implications to our society} 
\label{subsec:implications}

While we are not yet at the stage of achieving AGI, there are significant technical and ethical challenges that must be addressed before reaching that milestone. Current GPAIS offerings already provide numerous functionalities, but they also come with certain risks. The ability of these systems to process vast amounts of data and make automated decisions raises concerns regarding privacy, ethics, and fairness in their implementation.

Under the umbrella term of \textit{Trustworthy AI}, recent research efforts in AI are focused on developing techniques and frameworks that enhance oversight, transparency, interpretability, and robustness, among others, \revision{with the aim of designing responsible AI systems \cite{DiazRodriguez2023}}. Those efforts become even more needed when developing GPAIS. By gaining a deeper understanding of the underlying mechanisms and limitations of models qualifying as such, we can refine the design and training processes to minimise the occurrence of hallucinations, and ultimately maximise the emergence of desirable and reliable behaviors in GPAIS. This ongoing exploration and mitigation of emergent abilities and hallucinations will contribute to the responsible development and deployment of GPAIS in the future. 

Another important aspect that must be taken into consideration in GPAIS is their sustainability. The training process of current GPAIS like ChatGPT exhibits an unprecedented demand for computing resources \cite{Schwartz2020}. Future GPAIS generations may imply even higher environmental impact in terms of carbon footprint \cite{Strubell2019}. This is calling for more sustainable and greener approaches  \cite{verdecchia2023systematic} that reduce the computational cost of GPAIS, before they can be massively adopted.

One of the main challenges in this regard continues to be the evaluation of the results and explanations of an AI model, which has been a long-lasting issue. The community should continue re-thinking these issues \cite{Burnell2023}, as these metrics are the only means to determine secure and effective use. Evaluating and explaining the results of a GPAIS, particularly in the open-world setting \cite{Jitendra23}, may prove to become a very challenging task, especially when they combine multiple (potentially black-box) AI models and use very broad sources of information. Nevertheless, generative AI models may help generate better interpretations, explanations, and reasoning over them \cite{Barredo2020,Ali2023}. 

\revision{In summary, trustworthy AI technologies, including human oversight, transparency, interpretability, and robustness, among others, must contribute to the effective management of AI risks and address emerging GPAIS prospects, ultimately reducing the uncertainty and concerns of the society about the use of these modern AI systems.} 

\subsection{Regulation and governance in GPAIS}
\label{subsec:regulation}

\revision{Vivid discussions on the regulation of GPAIS, foundation models, and GenAI are taking place within the community as advances in GPAIS  are continually taking place.} While regulatory efforts for closed-world GPAIS may be more manageable, the open-world setting poses unresolved difficulties, as the specific tasks to be tackled can be largely unknown while the model is audited. This uncertainty makes it difficult to anticipate potential outcomes, even with valid metrics in place. The definition of GPAIS proposed in \cite{Campos2023} aimed to limit the set of AI models that would class as GPAIS, to those that may display emergent abilities, as they are more likely to inflict risks. Nevertheless, there are other challenges in the open-world setting, such as hallucinations or untrue statements/outputs, which call for an external audit process that would help us trust the behaviour of such an autonomous system.

The recent advancements in GPAIS have raised concerns regarding the need for prompt regulatory measures to ensure their safe deployment in society  (See Section 6.5 in \cite{DiazRodriguez2023} for a brief discussion). However, different countries are taking divergent approaches in this regard. 

The EU AI Act\footnote{\url{https://artificialintelligenceact.eu/} (Last access: 2023/10/28)} envisions a distinct regulatory framework compared to the proposals under consideration in the United Kingdom. In \cite{Bommasani2023}, an analysis was conducted to assess the compliance of the latest LLMs with the proposed EU AI Act, revealing their non-compliance. Developing standards to support the AI Act or any other regulatory framework will be faced with the task of specifying the current best practices in trustworthy and responsible AI \cite{Hupont2023}.

\revision{The recent EU AI Act discussion on GPAIS and foundations models\footnote{\url{https://www.europarl.europa.eu/news/en/press-room/20230505IPR84904/ai-act-a-step-closer-to-the-first-rules-on-artificial-intelligence} (Last access: 2023/10/28)} considers that they should guarantee robust protection of fundamental rights, health and safety and the environment, democracy and rule of law. They should assess and mitigate risks, comply with design, information and environmental requirements and register in the EU database. Generative foundation models, like GPT, should comply with additional transparency requirements, such as disclosing that the content was generated by AI, designing the model to prevent it from generating illegal content, and publishing summaries of copyrighted data used for training.}

\revision{Regulation is always associated to the 
auditability and accountability during its design, development, and use, according to specifications and the applicable regulation of the domain of practice in which the AI system is to be used, to design a responsible AI system \cite{DiazRodriguez2023}. Auditability refers to a property sought for the AI-based system, which may require transparency (e.g., explainability methods, traceability), measures to guarantee technical robustness, etc. The auditability of a responsible AI system may not necessarily cover all requirements for trustworthy AI, but rather those foretold by ethics, regulation, specifications and protocol testing adapted to the application sector (i.e., vertical regulation).}

Accountability is another crucial aspect to consider in regulation. Determining who is responsible for the outputs of a GPAIS, particularly in cases of unexpected emergent abilities, becomes essential when decisions made thereof bring about fatal consequences. From a scientific point of view, understanding the principles of design of these techniques, how they can be built, and their properties and limitations can help prescribe the regulatory directives that should be put in place for GPAIS. Even in the current developmental status of GPAIS, many ethical and legal aspects of their practical use remain without consensus, providing ample space for further debate.

\revision{
A recent ``Policy Brief'' on AI risk management standards for GPAIS and foundation models has been adopted by the UC Berkeley Center for Long-Term Cybersecurity (CLTC) \footnote{\url{https://cltc.berkeley.edu/publication/policy-brief-on-ai-risk-management-standards-for-general-purpose-ai-systems-gpais-and-foundation-models/} (Last access: 2023/10/28)}. It highlights key policy implications of the profile, as well as the considerations } \revision{of what AI risk-related policies would be especially valuable beyond the profile. They recommend employing the following three strategies as they seek to regulate GPAIS, foundations models, and GenAI.}
\revision{
\begin{enumerate}
    \item ``Ensure that developers of GPAIS, foundation models, and generative AI adhere to appropriate AI risk management standards and guidance.
 \item Ensure that GPAIS, foundation models, and generative AI undergo sufficient pre-release evaluations to identify and mitigate risks of severe harm, including for open source or downloadable releases of models that cannot be made unavailable after release. 
 \item Ensure that AI regulations and enforcement agencies provide sufficient oversight and penalties for non-compliance.''
\end{enumerate}

These discussions are directly linked to the importance of managing AI risks, as it is discussed in \cite{bengio2023managing}\footnote{\url{https://managing-ai-risks.com/} (Last access: 2023/10/28)}: ``We must anticipate the amplification of ongoing harms, as well as novel risks, and prepare for the largest risks well before they materialize. Climate change has taken decades to be acknowledged and confronted; for AI, decades could be too long.''

Together with the discussion on the societal-scale risk and the technical developments and to advance toward trustworthy AI technology to develop responsible AI systems, another important aspect is pointed out in \cite{PALLADINO2023,Almeida2023} and other recent publications on governance and the need for governance measures: ``For AI systems with hazardous capabilities, we need a combination of governance mechanisms matched to the magnitude of their risks.  Regulators should create national and international safety standards that depend on model capabilities. They should also hold frontier AI developers and owners legally accountable for harms from their models that can be reasonably foreseen and prevented. These measures can prevent harm and create much-needed incentives to invest in safety. Further measures are needed for exceptionally capable future AI systems, such as models that could circumvent human control.'' 

Paying attention to these necessary measures, we highlight a report published by the UC Berkeley Center for Long-Term Cybersecurity (CLTC) \footnote{\url{https://cltc.berkeley.edu/publication/a-taxonomy-of-trustworthiness-for-artificial-intelligence/} (Last access: 2023/10/28)} that aims to help organisations develop and deploy more trustworthy AI technologies, including 150 properties related to one of seven ``characteristics of trustworthiness'' as defined in the NIST AI RMF \footnote{\url{https://www.nist.gov/itl/ai-risk-management-framework} (Last access: 2023/10/28)}: valid and reliable, safe, secure and resilient, accountable and transparent, explainable and interpretable, privacy-enhanced, and fair with harmful biases managed. Using these characteristics as a starting point, the CLTC report names 150 properties of trustworthiness, which are mapped to particular parts of the AI lifecycle where they are likely to be particularly critical. Each property is also mapped to specific parts of the AI RMF core, guiding readers to the sections of the NIST framework that offer the most relevant resources.}

\revision{In summary, governance frameworks together with regulation and trustworthiness technologies must be developed cooperatively to ensure that an AI system in general -- and a GPAIS in particular --  is designed and engineered to achieve its goals, while: a) maintaining the ability to disengage or deactivate the system if necessary; and b) ensuring that an AI system would not have incentives to resist or deceive its operators. In a nutshell, they must be developed cooperatively for the safe deployment of GPAIS as responsible AI systems.}

\section{Conclusions} \label{sec:conclusions}

Many researchers are working on GPAIS, both to design new GPAIS and to define what they are. The principal goals of this work have been two-fold: (1) proposing a more comprehensive definition of GPAIS with a focus on their properties and functionalities, and (2) categorising different approaches to build them. In comparison with existing alternatives, our proposed definition allows for a more general view of GPAIS, considering different degrees of autonomy and expected capabilities. Together with these two principal goals, we have analyzed shortly GenAI as the most important foundation models and the multimodality as a crucial aspect for managing multiple inputs. Finally, we have discussed the prospects posed by GPAIS, the implications to our society, and the regulation and governance of these emerging systems.

There are a multitude of approaches to making an AI system more general. In order to consolidate the most relevant ones in a cohesive manner, we have proposed a taxonomy of methods. This taxonomy conceptually distinguishes between AI models that rely on other AI models to achieve generalisation abilities and those that utilise a single AI model. More classical multi-task learning approaches and foundation models have been categorised as AI systems in which a single AI model exists, while alternative approaches may use two or more AI systems, what we called AI-powered AI to introduce generalisation capabilities.  

The field of GPAIS is continually evolving, and the proposed taxonomy has established a robust foundation for understanding the diverse existing approaches. While LLMs are currently in the spotlight, there is a broad spectrum of approaches that can significantly contribute to the realisation of GPAIS. However, technical and ethical challenges must be necessarily discussed and addressed before stepping towards AGI. In the meantime, we must be mindful \revision{in the arrival of new open-world GPAIS models and task, on managing AI risks associated with current GPAIS, and work responsibly towards anticipating and mitigating such risks effectively via regulation and governance}.

\section*{Acknowledgments}

I. Triguero is funded by a Maria Zambrano Senior Fellowship at the University of Granada. I. Triguero, F. Herrera, D. Molina, and J. Poyatos are supported by the R\&D and Innovation project with reference PID2020-119478GB-I00 granted by Spain's Ministry of Science and Innovation and European Regional Development Fund (ERDF). J. Del Ser would like to thank the Basque Government for the funding support received through the EMAITEK and ELKARTEK programs (ref. KK-2023/00012), as well as the Consolidated Research Group MATHMODE (IT1456-22) granted by the Department of Education of this institution.

\bibliographystyle{unsrt}  
\bibliography{references}

\end{document}